\documentclass[10pt,twocolumn,a4paper]{esaAI}

\title{Spiking monocular event based 6D pose estimation for space application}

\def\authorEmail{jonathan.courtois@univ-cotedazur.fr}

\author[1]{Jonathan Courtois\thanks{Corresponding author. E-Mail: \authorEmail}}
\author[1]{Benoit Miramond}
\author[1]{Alain Pegatoquet}
\affil[1]{\textit{LEAT, Univ. Côte d'azur, CNRS, France}}

\begin{document}

\makeCustomtitle

\begin{abstract}
With the growing interest in on On-orbit servicing (OOS) and Active Debris Removal (ADR) missions, 
spacecraft poses estimation algorithms are being developed using deep learning to improve the precision of this complex task 
and find the most efficient solution. 
With the advances of bio-inspired low-power solutions, such as spiking neural networks and event-based processing and cameras, 
and their recent work for space applications, 
we propose to investigate the feasibility of a fully event-based solution to improve event-based pose estimation for spacecraft. 
In this paper, we address the first event-based dataset SEENIC with real event frames captured by an event-based camera on a testbed. 
We show the methods and results of the first event-based solution for this use case, 
where our small spiking end-to-end network (S2E2) solution achieves interesting results over $21cm$ position error and $14^\circ$ rotation error, 
which is the first step towards fully event-based processing for embedded spacecraft pose estimation.
\end{abstract}

\section{Introduction}

In recent years, the democratization of access to space has led to an unprecedented increase in spacecraft launches and large-scale constellation projects.
As a result, the orbits around our planet are becoming congested and the risk of collisions is increasing due to the presence of fast-moving space debris \cite{esareport}.
Recognizing the potential dangers, the Inter-Agency Space Debris Coordination Committee (IADC) has established guidelines for the containment of space debris and the safe disposal of satellites at the end of their operational life.
Agencies and companies have planned missions such as On-Orbit Servicing (OOS) or Active Debris Removal (ADR) \cite{grumman}\cite{Dubanchet2021}\cite{biesbroek2021clearspace} to extend the life of satellites and address the problem of space debris.
OOS missions provide services such as refueling, repairs and even the removal of end-of-life satellites from orbit. A critical component of this is the ability to accurately determine the attitude - translation and rotation - of spacecraft, especially when dealing with non-cooperative targets that cannot provide sensor data for attitude determination.
The challenge is exacerbated by the dynamic and complex space environment, where factors such as changing lighting and the small size of distant spacecraft targets increase the difficulty of pose estimation\cite{paulyleo}.
As resources on board spacecraft are limited, pose estimation for Guidance Navigation and Control (GNC) must not only be accurate, but also energy and computationally efficient.
Vision-based systems using LIDARs and RADARs offer advantages in terms of accuracy and are therefore the first choice for spacecraft attitude determination. However, the volume and power consumption of these systems can be a challenge, in addition to the deployment difficulties induced \cite{paulyleo}. With a monocular camera, the volume, power consumption and complexity of deployment could be reduced with the same or lower accuracy \cite{cassinis2022monocular}.
in the search for the most efficient solution, we propose to investigate the feasibility of such a system with emergent event-based solution by using event-based camera (EBC) and spiking neural networks (SNN).
These sensors and processing methods are already attracting growing interest in the space community \cite{izzo2022neuromorphic} with the first SNN on board in space \cite{abderrahmane2022spleat} and studies on EBC behaviour under radiation \cite{roffe2021neutron}.
With this paper, we propose the first fully event-based approach for spacecraft pose estimation, but also a novel method to account for the event stream.
Section 2 introduces the event-based camera, spiking neural network and pose estimation for space application. In Section 3 we present the dataset and the network used, and finally in Section 4 we discuss the results and future works.

\section{Related Work} \label{Related work}
    
\subsection{Event-based camera} 
The retina-inspired EBCs are sensors that respond to fluctuations in light intensity with a high temporal resolution ( $\sim 1\mu s$).
Each pixel of the grid detects its change and discretizes it into positive and negative events over time.
this behavior leads to a pixel-independent output of the event stream, which enables a high dynamic range (HDR) through the grid \cite{Gallego2019}.
Consequently, these cameras do not output an image, but an event stream that contains the position of the pixel, the polarity of the event and the associated timestamp (x,y,p,t).
Different representations of this event stream are used, such as graphs \cite{Bi2020}, event frame reconstruction \cite{maqueda2018event}\cite{fang2020incorporating}\cite{cordone2021learning} or motion-compensated event-frame \cite{stoffregen2019event}.
There is a growing interest in EBC for space applications, both from Earth for Space Situation Awareness (SSA) \cite{Ralph2023} as well as in space for landing applications \cite{mcleod2022globally} and pose estimation \cite{jawaid2023seenic} \cite{rathinam2023spades}.
In \cite{roffe2021neutron} they investigate radiation effects for purely event-based computation and characterize the generated noise in terms of events uniformly distributed over the sensor's field of view, which makes their use in space interesting.
The use of spiking neural networks enables the treatment of the EBC output in an event-based manner.

\subsection{Spiking Neural Networks} 
SNNs are biologically inspired neural networks for deep learning applications that show interesting results in terms of energy consumption.
They consist of neurons that mimic the information transmission of neurons in our brain.
By accumulating input stimuli in their membrane potential over time, they emit an output impulse, a so-called spike, when their potential reaches a certain threshold.
Different variants of spiking neuron activation have been studied using the Integrate and Fire (IF) and Leaky Integrate and Fire (LIF) \cite{eshraghian2023training} models, which are mainly used in SNN.
During training, the temporal dynamics of spiking neurons are usually expressed with time steps that discretize time into a series of passes through the network, using the same or a different input for each pass, depending on the method used, which can also introduce latency issues.
LIF neurons introduce leak dynamics that reduce the membrane potential across these time steps and allow for better temporal dynamics.
When a spike is fired, spiking neurons reset their potential in two ways: with a hard reset, which means that the potential is set to a specific level (generally 0), or with a soft reset, where the value of the threshold is subtracted from the membrane potential \cite{eshraghian2023training}.

With a binary flow of information through the network, SNNs benefit from activation with higher sparsity and fewer multiplication operations than Formal Neural Networks (FNN), which can have an impact on final energy consumption.
Even though temporal memory capacity could entail more latency and memory accesses during inference, despite the membrane potential yet to be characterized, they could be up to 8 times more energy efficient than FNNs when running on dedicated hardware \cite{Lemaire2022} \cite{Dampfhoffer2022}.

SNNs are able to solve complex tasks in an event-based manner, such as automotive object detection \cite{Cordone2022} and even embedded \cite{courtois2024embedded}, and show interesting results for resource-constrained environments.
For the first time, SNN were used on board satellite to detect clouds in the ESA OPS-SAT mission \cite{abderrahmane2022spleat}.
Based on these results, in this paper we investigate the feasibility of using event-based flows for 6D monocular position estimation of spacecraft.

\subsection{Spacecraft poses estimation}
monocular pose estimation of spacecraft refers to the process of determining the position and orientation (pose) of a spacecraft in space based on a single image sensor.
In space rendezvous, OOS and ADR, the goal is to accurately estimate the relative pose (translation and rotation) of a target spacecraft with respect to a reference frame on the servicer spacecraft.
Despite promising results, deep learning (DL) methods for real missions are still very computationally intensive and still show a significant performance degradation when trained with synthetic data and testing with real images \cite{paulyleo}.
the hybrid modular approach and the direct end-to-end approach are the two best known methods in the literature (e.g. survey \cite{paulyleo}). While the hybrid modular approach tends to combine different DL methods and algorithms, such as object detection, keypoint estimation and pose computation, the direct end-to-end method consists of feeding a neural network with the input to output the poses estimation directly.
For each method, the threshold between implementation complexity and performance is drawn.
To measure this performance, the mean position and orientation errors between the prediction 
($trans_{pred},rot_{pred}$) and the ground truth ($trans_{gt},rot_{gt}$) are used. 
The mean position error is calculated using the Euclidean norm $E_{t}=||(trans_{pred}-trans_{gt})||_2$. 
The mean orientation error is calculated using the absolute dot product between the two rotation vectors 
with the following formula: $E_{r}=2*arcos(||rot_{pred}-rot_{gt}||)$.
Based on this metric and \cite{paulyleo}, we can largely include the translation error between $[0.005;1.192]m$, the rotation error between $[0.013,14.350]^\circ$ for models with almost one million parameters up to more than 200 parameters. 
This large overview applies to several datasets, but as far as we know, no one has yet tackled event-based pose estimation datasets with spiking neural networks.
So far, only one event-based spacecraft pose estimation datasets have been published, namely SEENIC \cite{jawaid2023seenic}.

\begin{figure*}[htbp]
    \centering
    \captionsetup{justification=centering}
    \includegraphics[width=.30\textwidth,angle=90]{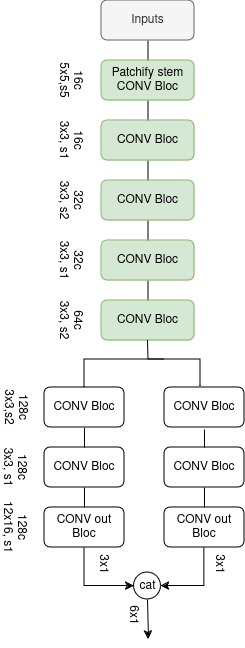}
     \caption{Small Spiking End to End network (S2E2) architecture. Each block give information about the channel (c) then kernels then stride (s).}
     \label{fig:s2e2}
 \end{figure*}

\section{Method}
\subsection{SEENIC Dataset}
Presented in \cite{jawaid2023seenic},SEENIC is the first spacecraft pose estimation dataset with synthetic and real event data.
Using the Hubble Space Telescope, 
SEENIC is proposed to evaluate the gap between synthetic V2E \cite{hu2021v2e} event frames and real event frames captured on a 
robotic testbench with an event-based camera.
although this dataset opens the possibility for event-based studies, it has some limitations.
For the training part, 
the dataset proposes a single trajectory consisting of a linear translation towards the target and a rotation around an axis. 
This scenario shows a lack of light variations and thus an unbalanced training set.
The event stream was generated using V2E software, which converts RGB images into an event stream.
This method has two limitations: First, the event stream is generated by an RGB sensor, 
which means that HDR and the high temporal benefit of the EVB sensor are lost. 
Second, the timestamp of the event stream is simulated around the timestamp of the RGB frame, 
resulting in unrepresentative temporal continuity of the event timestamp in the generated stream.
In addition, The proposed event stream shows only one polarity as opposed to the two polarities included in EBC.
The test set consists of 20 scenarios divided into 10 linear translation approaches and 10 orbiting inspection trajectories. 
In each of the 10 and 10 cases, these trajectories are the same but for 5 different illumination conditions under a slow and a fast approach.
This cut also leads to an unbalanced test set with a total duration of 33.55s for the fast scenario versus 512.28s 
for the slow scenario with a total of 63 million events and 325 ground truth poses versus 486.4 million events and 5090 ground truth poses.
In addition, the mock-up used was a 3D-printed version of Hubble, 
which did not reflect the actual optical properties of the satellite material. 
Since the EBC camera produces events with a high temporal resolution, 
the frequency of the light becomes visible when the event stream is examined at a slower speed than the poses.
Furthermore, the ground truth poses, for the trainset, are given in the form of a 3D vector for position and a rotation matrix for rotation, 
but in the test the poses are represented as a 6D vector (x,y,z,Rx,Ry,Rz) where the rotation angle appears to be given in radians.

To get as close as possible to a real scenario and avoid domain gap between test and train set, 
we decide to treat this dataset with our own approach, explained in more detail in section 4.
\begin{figure*}[btp]
    \centering
    \captionsetup{justification=centering}
    \includegraphics[width=.9\textwidth]{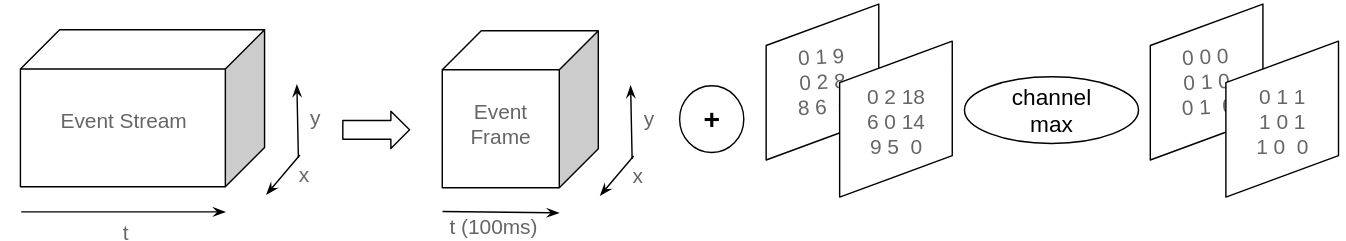}
     \caption{Reconstruction of the Event Frames from the Event stream to enable GPU training}
     \label{fig:streamevent}
 \end{figure*}

\subsection{Network}
To predict the 6D positions from the event images with a direct end-to-end approach, 
we decide to create a network inspired by the Spacecraft Pose Network (SPN) \cite{sharma2019pose} and the 
Small 32 ST-VGG \cite{Cordone2022}.
The Small Spiking End to End Network (S2E2) \ref{fig:s2e2} is a small network with 0.625 million parameters.
Like the Small 32 ST-VGG, it starts with a patchify stem convolutional block used to reduce the input dimension, 
and then 4 convolutional blocks that feed two paths of 3 convolutional blocks each, 
each ending with a convolutional output block. 
Each path is used to predict either the position (x,y,z) or the rotation (rx,ry,rz). 
The convolutional output blocks are the same as the regular blocks, but without spiking neurons to allow for floating point output.
These two predictions are then simply combined to obtain our 6D estimate.
Each convolutional block consists of a convolutional layer, a batch normalization (BN) layer and an activation neuron.
To have better comparison material, 
we trained 6 versions of the S2E2 network. 
2 formal versions are with Relu activation (with and without BN) as baseline, 
the other versions have parametric LIF (PLIF) neurons as activation (with and without BN) and compare a StepLr scheduler and a CosineAnnealingLR.
PLIF neurons are spiking neurons that have a trainable decay factor.
In the formal version of the network, the BN is placed after the convolutional layer, but in the Spiking ones, 
the BN is placed before the convolutional layer. 
The choice of scheduler and the placement of the BN in the CONV 
block are similar to the 32-ST-VGG in \cite{cordone2022performance} for performance reasons.
Even though the BN introduces a multiplication operation into the network and interrupts 
the dynamics of the event flow in the network, 
it can be merged with the convolution during inference to restore the dynamics of spiking information transmission.
 
\section{Experiments}
We used our SNN model in a direct end-to-end approach with the SEENIC dataset.
To train our models we use the pytroch litghning framework and the spikingjelly framework \cite{fang2023spikingjelly} 
for the PLIF models.
 
\begin{table*}[h]\renewcommand{\arraystretch}{1}
    \centering 
    \captionsetup{justification=centering} 
    \caption{Comparison of performances between the different versions of our 625k parameter network. The results are the average of 3 K-fold runs followed by the [min;max] value of the K-folds}
    \begin{tabular}{ c || c | c } 
        \hline\hline
        Model & Mean Position Error ($E_t$) (m) & Mean Rotation Error ($E_r$) ($^\circ$)\\
        \hline\hline
        Relu\hspace{1.1cm}W/O BN                & $0.14\ [0.07;0.22]$   & $\ 8.4\ [5.7;12.5]$\\
        PLIF Steplr W/O BN                      & $0.31\ [0.14;0.44]$   & $17.2\ [8.8;24.1]$\\
        PLIF\hspace{2mm}Coslr  W/O BN           & $0.24\ [0.12;0.38]$   & $13.8\ [4.0;23.9]$\\ 
        Relu\hspace{1.1cm}W\hspace{4mm}BN       & $0.09\ [0.04;0.16]$   & $\ 7.6\ [3.6;11.6]$\\ 
        PLIF Steplr W\hspace{4mm}BN             & $0.23\ [0.14;0.35]$   & $16.1\ [9.2;25.8]$\\
        PLIF\hspace{2mm}Coslr  W\hspace{4mm}BN  & $0.21\ [0.10;0.35]$   & $14.3\ [7.1:23.4]$\\ 
        \hline\hline
    \end{tabular}
    \label{tab:results}
\end{table*}

\subsection{Custom Dataset}
To avoid some limitations of the dataset, domain gap between test and training data, and to work with real event data, 
we use a custom split of the training and test set based on the original test set of the SEENIC dataset.
We decided to split the 5415 binary frames into 80\% for the training set and 20\% for the test set using an algorithm.
This handcrafted algorithm splits the data sets so that the data loader receives a sequence of 10 consecutive event 
frames when a frame index is selected for training.
This method allows the network to work with sequences instead of single frames and thus train the temporal dynamics 
of the spiking neurons with a soft reset.
To approximate the frequency of poses to ground truth, we moulded the event stream into a series of 100ms 
event frames in which we accumulated the event count per pixel in both channels according to polarity.
Then, for each pixel, we decided to keep an event on the strongest channel or no event if there is none or 
both are equal \ref{fig:streamevent}.
In this way, we obtained 5415 binary frames with 2*480*640 pixels labelled with a 6D pose.
Since we touched the split of the dataset, we repeated our experiment 3 times with a K-fold cross-validation 
to get a better idea of the performance of our network.
In addition, two types of data augmentations were used, such as random event noise with a probability 
of $10\%$ and $10\%$
probability of a dead pixel during a sequence (spikes are ignored by this pixel).

\subsection{Training parameters}
The training of the S2E2 networks was performed with short sequences of 10 consecutive binary images, 
with a prediction and a loss calculation for each image as it had one timestep.
the loss used corresponds to the sum of the Euclidean norms of the position and rotation error vectors. 
Since the poses of the ground truth are specified with the position in meters and the rotation in radians, 
the following formula results:
\begin{equation}
 loss = ||(trans_{pred}-trans_{gt})||_2 + ||(rot_{pred}-rot_{gt})*\frac{180}{\pi}||_2
\end{equation}
A batch size of 100 over 100 epochs was used. The learning rate was set to $1.10^{-3}$ 
with a StepLr scheduler with the following parameters 
$step\_size = 10; gamma = 0.5$, for the formal network and the spiking version if it is precised, 
or with a CosineAnnealingLR scheduler with these parameters $T_{max} = 100; eta_{min}=1.10^{-6}$.
   
\section{Results and Discussion}

If we compare the different results, we see that our S2E2 network with the CosineAnnealingLR scheduler achieves an average error of $21.2cm$ and $14.3^\circ$ while the best k-fold set achieves an error of
$10.6cm$ and $7.1^\circ$.
Our FNN baseline, which achieves an average error of $9.1cm$ and $7.6^\circ$,
with the best k-fold set achieving an error of $4.0cm$ and $3.6^\circ$.
Even though our SNN version has a larger error than the FNN version, it still manages to converge to an acceptable solution in the isoepoch comparison.
The traning results show that the FNN baseline seems to have reached a stable level after 100 epochs, while the SNNs version is not stabilised and a larger number of epochs must be used.
The effect of the BN in this case contributes to instability in training, but helps to converge to a better estimation.
For classification tasks, CosineAnnealingLR also produces better results than the StepLr scheduler.
And although all our networks converge to a solution, there is an average gap of -$12.6cm$ and $27.7cm$, $7.1^\circ$ and $19.4^\circ$- between the best and the worst k-fold set for all versions of the FNN and SNN networks respectively.
This difference shows the imbalance of the data set and the need for a larger data set to achieve better generalisation and then comparison.
Even though these results are from a small use case, they show the feasibility of event-based processing from the sensors to the network and open the door to an event-based processing pipeline in space, which the SNN low-power solution also brings.
To enable a better and fairer comparison with the formal approach, others event-based datasets are needed for space applications.
Recent work shows a second event-based dataset in procedings with a larger number of scenarios called SPADES \cite{rathinam2023spades},
which we will address in future work.
Also, the usability of hybrid solutions and addressing the domain gap between synthetic datasets and real event datasets, such as key point prediction, could provide some interesting features in an event-based scenario.
Finally, testing the pipeline on neuromorphic dedicated hardware with measurement of latency and energy consumption could bring us closer to a real onboard solution.

\section{Acknowledgement}
This material is based upon work supported by the French national research agency (project DEEPSEE ANR-17-CE24-0036) 
and by the donation from Franck DIARD to the Universite Cote d'Azur.

\printbibliography
\addcontentsline{toc}{section}{References}
\end{document}